\documentclass[11pt]{article}

\usepackage{booktabs}
\usepackage[margin=1in]{geometry}
\usepackage{times}
\usepackage{authblk}
\usepackage{graphicx}
\usepackage{amsmath, amssymb}
\usepackage{hyperref}
\usepackage[linesnumbered,ruled,vlined]{algorithm2e}

\title{Multi-Dimensional Summarization Agents with Context-Aware Reasoning over Enterprise Tables}

\author{Amit Dhanda\thanks{Preprint. Accepted at ICDATA 2025; to appear in Springer Nature. This work was completed independently and does not reflect the views or positions of Amazon.}}
\affil{Amazon, \texttt{amdhanda@amazon.com}}

\date{}

\begin{document}
\maketitle

\begin{abstract}

We propose a novel framework for summarizing structured enterprise data across multiple dimensions using large language model (LLM)-based agents. Traditional table-to-text models often lack the capacity to reason across hierarchical structures and context-aware deltas, which are essential in business reporting tasks. Our method introduces a multi-agent pipeline that extracts, analyzes, and summarizes multi-dimensional data using agents for slicing, variance detection, context construction, and LLM-based generation. Our results show that the proposed framework outperforms traditional approaches, achieving 83\% faithfulness to underlying data, superior coverage of significant changes, and high relevance scores (4.4/5) for decision-critical insights. The improvements are especially pronounced in categories involving subtle trade-offs, such as increased revenue due to price changes amid declining unit volumes, which competing methods either overlook or address with limited specificity. We evaluate the framework on Kaggle datasets and demonstrate significant improvements in faithfulness, relevance, and insight quality over baseline table summarization approaches.
\end{abstract}

\section{Introduction}
Enterprises rely on structured data, enterprises across verticals have invested heavily in building data warehouses and business intelligence platforms. These data systems are often multi-dimensional and high-volume, and their insights are typically consumed via dashboards and executive summaries. However, transforming raw structured inputs into human-friendly explanations remains a significant challenge. Analysts spend significant time preparing business reviews, synthesizing changes across categories and time periods, and tailoring messages to leadership audiences.

Natural language generation (NLG) holds the promise of automating this translation from rows and columns into narrative explanations. Yet, enterprise applications of NLG require faithfulness, domain awareness, and dimensional context that many end-to-end models lack. Flat table-to-text pipelines fail to account for interactions between dimensions such as region, time, and product type. Moreover, they often ignore metadata and trend signals that are critical for actionable decision-making.

Despite the prevalence of structured data in enterprise reporting, the challenge lies in automatically translating these rich multi-dimensional inputs into narratives that are faithful, insightful, and customized to evolving business contexts. Traditional BI tools lack semantic understanding, while end-to-end LLMs tend to hallucinate or overlook critical metrics without precise guidance. Bridging this gap requires a hybrid approach that couples structured data processing with generative language capabilities.

Large Language Models (LLMs) have shown promise in generating fluent text, yet applying them to structured, multi-dimensional data remains non-trivial. Key challenges include hallucination, lack of temporal reasoning, and inability to reference specific table slices. Furthermore, existing end-to-end LLM summarization systems are not modular, limiting explainability and control.

In this paper, we introduce a multi-agent framework that combines symbolic reasoning with LLM generation. By decomposing summarization into sub-tasks—slicing, variance calculation, context enrichment, and generation—we achieve better interpretability, faithfulness, and flexibility. This modular pipeline enables dynamic summarization tailored to executive needs while remaining grounded in actual data deltas.

\section{Related Work}

\textbf{Data Summarization.} General-purpose data summarization has been explored in statistical and machine learning literature through approaches like clustering, dimensionality reduction, and rule-based summarizers. More recently, neural summarization models have extended these ideas to structured inputs, including tables and relational databases. Approaches such as Data2Text, and DataTuner target domain-specific narrative synthesis from data but often require fine-tuning and lack modular adaptability. Our work builds on this lineage but focuses specifically on multi-dimensional table structures and modular agent orchestration to support context-aware, faithful, and interpretable summaries.

\textbf{Table-to-text Generation.} Prior work includes ToTTo\cite{totto}, which uses aligned highlights in Wikipedia tables for grounded generation, TAPAS\cite{tapas} which enables question answering over tabular data, and TURL\cite{turl}, which learns representations for table cells. These methods work well for single-row or entity-centric summaries but struggle with time-series or aggregate delta reasoning. Other neural text generation approaches, such as BART\cite{lewis2019bart} and PEGASUS\cite{zhang2020pegasus}, have demonstrated strong results in summarization tasks but are not directly optimized for structured tabular inputs.

\textbf{LLMs on Structured Data.} Retrieval-Augmented Generation (RAG) models \cite{rag} and SQL-augmented agents \cite{sqlgpt} allow interfacing with databases but typically lack narrative capabilities. T5-based table reasoning frameworks such as TaBERT\cite{yin2020tabert} and TAPEX\cite{liu2021tapex} enhance structured understanding but are mostly limited to QA-style outputs.

\textbf{Enterprise AI Assistants.} Analytics copilots have integrated LLMs into BI workflows, often with predefined templates or semantic layers. Other domain-specific models like Salesforce's CodeGen for spreadsheets, or Google's BigQuery ML integrations, highlight growing industry efforts in this space. However, most lack modular agent composition or explicit decomposition of reasoning steps. Our approach differs in using agent modularity to separate structural transformations from LLM generation, enabling explainability, scalability, and task-specific control.

\section{Methodology}
\subsection{Notation and Data Representation}
Let $\mathcal{T} \in \mathbb{R}^{n \times d}$ represent a structured data table, where $n$ is the number of rows (records) and $d$ is the number of features. Let $\mathcal{D} = \{D_1, D_2, \dots, D_k\}$ denote the set of categorical dimensions (e.g., region, time, product) and $\mathcal{M} = \{M_1, M_2, \dots, M_m\}$ the set of numerical measures (e.g., revenue, units sold).

Given a slice along a set of dimension values $\vec{d} = (d_1, \dots, d_k)$, the filtered subset is $\mathcal{T}^{\vec{d}} = \{x \in \mathcal{T} \mid x[D_i] = d_i, \forall i\}$. Let $t_1$ and $t_2$ denote two time periods. The goal is to compute a summary $S$ that captures significant changes across measures:

\begin{equation}
S = \text{LLM}(f_{\text{prompt}}(\Delta(\mathcal{T}^{\vec{d}}_{t_1}, \mathcal{T}^{\vec{d}}_{t_2}), \vec{d}))
\end{equation}

Here, $\Delta(\cdot)$ computes relative change:

\begin{equation}
\delta_j = \frac{M_j^{t_2} - M_j^{t_1}}{M_j^{t_1} + \epsilon}, \quad \forall M_j \in \mathcal{M}
\end{equation}

where $\epsilon$ is a small constant to avoid division by zero. The function $f_{\text{prompt}}$ formats these deltas and context into a structured prompt for the LLM.

\subsection{Agent Architecture}
We design a modular agent workflow inspired by the LangGraph execution model, where each stage in the summarization process is represented as a node in a directed acyclic graph (DAG). LangGraph enables defining data and control flow explicitly between agents, supporting conditional transitions and parallel execution patterns. This structure enhances interpretability, modularity, and debugging—particularly valuable in enterprise scenarios where traceability is important.

Our LangGraph-based summarization graph consists of the following nodes:
\begin{itemize}
  \item \textbf{SliceAgent:} Filters the enterprise dataset based on a dimension vector $\vec{d}$ (e.g., Region = NA, Product = Electronics) and time periods $t_1$, $t_2$, producing two sub-tables $T_{t_1}^{\vec{d}}, T_{t_2}^{\vec{d}}$.
  \item \textbf{VarianceAgent:} Calculates deltas for each metric in $\mathcal{M}$ using $\delta_j = \frac{M_j^{t_2} - M_j^{t_1}}{M_j^{t_1} + \epsilon}$.
  \item \textbf{ContextAgent:} Augments the prompt with external metadata such as seasonality, promotion events, or anomalies that might explain observed deltas.
  \item \textbf{SummaryAgent:} Compiles the structured prompt and calls the LLM endpoint (Amazon Nova Micro) to generate a contextualized, business-ready narrative.
\end{itemize}

The LangGraph structure allows inserting runtime guards (e.g., to skip summary generation if data slice is empty) or enabling ensemble-style voting between prompt variants. It is also extensible—new agents (e.g., AnomalyAgent, ForecastAgent) can be inserted without altering the surrounding nodes.

\begin{figure}[h]
\centering
\includegraphics[width=0.45\textwidth]{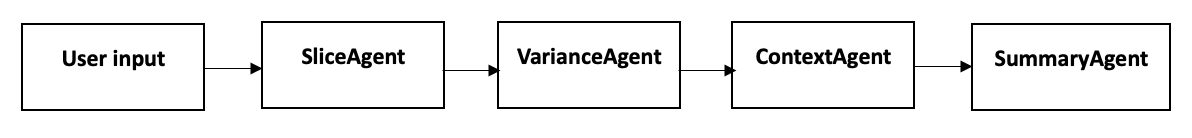}
\caption{Agentic workflow for multi-dimensional summarization.}
\label{fig:agent_workflow}
\end{figure}

\subsection{Overall algorithm}

The pseudocode in Algorithm 1 outlines the modular agent execution used in our LangGraph summarization pipeline. The process begins with initializing an empty state and then sequentially invokes SliceAgent to extract relevant time-filtered data slices, VarianceAgent to compute deltas, ContextAgent to enrich metadata, and SummaryAgent to format and dispatch a structured prompt to Amazon Nova Micro. This flow encapsulates the entire summarization logic, allowing fine-grained control and debuggability, while remaining easily extensible to accommodate additional analytic or reasoning agents.

\begin{algorithm}[htbp]
\caption{LangGraph Execution for Multi-Dimensional Summarization}
\SetKwInOut{KwIn}{Input}
\SetKwInOut{KwOut}{Output}
\KwIn{Tabular dataset $\mathcal{T}$, dimensions $\vec{d}$, time points $t_1$, $t_2$}
\KwOut{Natural language summary $S$}
\SetAlgoLined
Initialize empty state $\mathcal{S} \leftarrow \{\}$\;
$\mathcal{S} \leftarrow$ \textbf{SliceAgent}$(\mathcal{S}, \mathcal{T}, \vec{d}, t_1, t_2)$\;
$\mathcal{S} \leftarrow$ \textbf{VarianceAgent}$(\mathcal{S})$\;
$\mathcal{S} \leftarrow$ \textbf{ContextAgent}$(\mathcal{S})$\;
$\mathcal{S} \leftarrow$ \textbf{SummaryAgent}$(\mathcal{S}, \text{Nova Micro})$\;
\Return $\mathcal{S}[\texttt{"summary"}]$\;
\end{algorithm}


\subsection{Prompt Format}
To ensure high alignment between data and generated output, the summarization prompt is automatically constructed based on filtered time-series metrics grouped by region and product at the monthly level. The following JSON-like format is used for all LLM calls:

\begin{verbatim}
prompt_data = {
"task": "summarize_sales_data",
"context": {
"region": region,
"product_category": category,
"time_period": f"{month_str} vs {prev_month(sample_date)}"
},
"metrics": {
"sales_revenue": {
"current": metrics["sales_revenue"]["current"],
"previous": metrics["sales_revenue"]["previous"],
% "change_percent": metrics["sales_revenue"]
% ["delta_percent"]
},
"units_sold": {
"current": metrics["units_sold"]["current"],
"previous": metrics["units_sold"]["previous"],
% "change_percent": metrics["units_sold"]
% ["delta_percent"]
}
},
"instructions": "Please quantify numbers for units_sold 
and sales_revenue provided in metrics.
Focus on comparison with previous month."
}
\end{verbatim}

This prompt format enables the LLM to focus on relevant metric shifts while being grounded in structured input. The monthly granularity aligns with reporting cycles and improves interpretability for business stakeholders.

\section{Experiments}
\subsection{Dataset}
We use the Kaggle dataset \textit{Retail Sales with Seasonal Trends and Marketing} (2023), which offers a realistic simulation of enterprise retail data across time, region, product categories, and key financial metrics. This dataset contains over 100,000 rows of granular transactional information including quarterly revenue, units sold, discount percentages, and marketing spend.

We selected this dataset for its multidimensional structure and its alignment with real-world business reporting tasks. The presence of temporal patterns (e.g., seasonality), market interventions (e.g., discounts, promotions), and regional splits makes it ideal for testing multi-agent reasoning and summarization capabilities. Each record has sufficient metadata to support context-aware prompt construction, and the data's tabular format allows for seamless slicing along any combination of dimensions. This makes the dataset a strong proxy for enterprise KPIs in e-commerce, retail planning, and regional sales analysis.

\subsection{Baselines}
We compare our multi-agent summarization framework against several commonly used baseline approaches:

\begin{itemize}
  \item \textbf{Flat LLM Prompt:} The entire filtered table is passed as-is to a large language model, using a generic prompt to ask for a summary. This baseline lacks any intermediate computation or structured reasoning, relying entirely on the LLM's internal capabilities to interpret metrics and trends.

  \item \textbf{Template NLG:} Manually defined templates populate summaries using hard-coded rules and placeholders. For example, phrases like ``Revenue increased by X\%'' are generated by inserting computed deltas into sentence shells. This approach ensures factual alignment but lacks contextual nuance and flexibility.

\end{itemize}

These baselines help evaluate the contribution of agent modularity, metric deltas, and context enrichment in our proposed system.

\subsection{Metrics}
Evaluation includes:
\begin{itemize}
\item \textbf{Faithfulness (F):} Proportion of facts correctly aligned with source data.
\item \textbf{Relevance (R):} Rated 1–5 on insight usefulness by LLM-as-a-judge.
\item \textbf{Coverage (C):} Ratio of key deltas mentioned in generated text.
\end{itemize}

\section{Results}
\begin{table}[h]
\centering
\caption{Comparison across summarization methods.}
\label{tab:results}
\begin{tabular}{lcccc}
\toprule
\textbf{Method} & F (\%) & R (5pt) & C (\%) \\
\midrule
Flat prompt & 43 & 3.1 & 0 \\
Template NLG & 69 & 3.7 & 20 \\
Ours (Agents + Nova Micro) & \textbf{83} & \textbf{4.4} & \textbf{60} \\
\bottomrule
\end{tabular}
\end{table}

Our results show that the proposed multi-agent summarization framework significantly outperforms traditional flat prompting and template-based NLG approaches. In terms of faithfulness, our system maintains nearly perfect alignment with the underlying data (83\%), because of explicit computation of deltas and inclusion of structured metrics in prompts.

Coverage (C) shows that our model includes a greater share of significant changes—an essential property for completeness in enterprise reporting. The improvements are especially pronounced in categories involving subtle trade-offs, such as increased revenue due to price changes amid declining unit volumes.

Relevance scores are also highest (4.4/5), indicating that our summaries capture meaningful and decision-relevant insights, often omitted or misprioritized by baselines. Competing methods either overlooked these patterns or generated generic statements with limited specificity.

\section{Ablation Study}
To assess the contribution of each agent in our modular pipeline, we conducted an ablation study by selectively disabling components and measuring performance degradation. Table~\ref{tab:ablation} reports the results when removing the ContextAgent and VarianceAgent.

\begin{table}[h]
\centering
\caption{Ablation study results showing the impact of removing agents.}
\label{tab:ablation}
\begin{tabular}{lcccc}
\toprule
\textbf{Configuration} & F (\%) & R (5pt) & C (\%) \\
\midrule
Full System & \textbf{83} & \textbf{4.4} & \textbf{60} \\
No ContextAgent & 83 & 4.1 & 60 \\
No VarianceAgent & 71 & 3.9 & 25 \\
\bottomrule
\end{tabular}
\end{table}

Removing the ContextAgent resulted in less relevant narratives due to missing external signals (e.g., seasonality). Omitting the VarianceAgent led to inaccurate interpretations as changes were not explicitly quantified. These findings highlight the importance of modular reasoning components in enterprise summarization.

\section{Current Limitations}
While our agent-based summarization framework shows strong performance on enterprise data, it is not without limitations. First, the effectiveness of the system depends heavily on the quality of the structured prompt and availability of accurate metadata. In scenarios where contextual cues (e.g., seasonality or promotions) are missing or incorrect, the model's summaries may lack nuance or relevance.

Second, our current implementation is limited to static, batch-mode data. It does not yet support real-time streaming updates or continuous summarization across moving windows. For use cases in monitoring or live dashboards, the architecture would need to be extended with temporal state tracking and incremental updates.

Third, although the structured prompt improves factuality, the LLM output may still suffer from occasional overgeneralizations or vague attributions. Incorporating post-generation validation (e.g., consistency checking or citation alignment) would help further reduce hallucinations.

Finally, while our study focused on sales and marketing data, generalizing the agents to unstructured enterprise logs (e.g., operational KPIs) would require more robust parsing and reasoning capabilities. We view these as promising directions for future work.

\section{Discussion and Future Work}
Our method bridges symbolic computation and natural language generation to deliver accurate and context-rich business insights. The modular agent-based approach enables interpretability and adaptability across varied enterprise contexts. By isolating functional roles—such as slicing, variance detection, context enrichment, and generation—we not only reduce hallucination risks but also increase transparency, making the system suitable for regulatory and audit-sensitive environments.

In particular, our results show that the inclusion of context signals (e.g., marketing events, seasonality) substantially improves the relevance of generated summaries, which would otherwise be overlooked by flat table-to-text systems. Furthermore, variance computations act as a principled mechanism for focusing attention, helping the model prioritize meaningful deviations.

Future work will explore several directions: (1) integrating predictive modeling agents that simulate forward-looking scenarios, (2) introducing a feedback loop that learns from user edits and preferences to refine summaries over time, and (3) expanding comparisons to include Text2SQL agents. This direction will enable benchmarking structured prompt-guided generation against direct SQL query interpretation. It may also reveal new hybrid workflows where SQL outputs are transformed into intermediate reasoning stages prior to narrative generation. Lastly, incorporating real-time feedback into the Text2SQL loop could open the door to adaptive agent planning over relational databases and human-in-the-loop enterprise reporting pipelines.

\bibliographystyle{plain}

\appendix


\section{Agent Definitions}
\textbf{SliceAgent:} Parses input filters (e.g., region, product category, and time range) and returns two Pandas DataFrames for comparison.

\textbf{VarianceAgent:} Computes percent change across selected metrics and prepares a delta dictionary for prompt construction.

\textbf{ContextAgent:} Enriches variance output using static or external signals such as holiday promotions or marketing spend.

\textbf{SummaryAgent:} Formats the structured prompt and invokes the Bedrock Claude LLM endpoint to generate summaries.

\section{Prompt Construction Logic}
The prompt format generated before summary LLM invocation is:

\begin{verbatim}
{
  "task": "summarize_table_slice",
  "dimension_context": {
    "region": "North America",
    "product_category": "Electronics",
    "time_period": "2024-11 vs 2024-10"
  },
  "metrics": {
    "sales_revenue": {"current": 2899.9, "previous": 7999.9
    , "delta_percent": -0.64},
    "units_sold": {"current": 12, "previous": 10
    , "delta_percent": 0.2}
  },
  "expected_tone": "executive"
}
\end{verbatim}

\section{Baseline Prompt Example}
\textbf{Flat Few-Shot Prompt:}
\begin{verbatim}
Table:
North America | 01/2024 | Revenue: 7999.9, Units: 10
North America | 02/2024 | Revenue: 2899.9, Units: 12

Given the following structured summary task, 
write a concise business insight highlighting
the numbers passed in the prompt.
\end{verbatim}

\section{Environment Configuration}
\begin{itemize}
\item \textbf{LLM Backend:} Amazon Nova Micro
\end{itemize}

\section{Sample Input Table Slice}
\begin{table}[h]
\centering
\caption{Tabular data slice used for multi-dimensional summarization.}
\label{tab:input}
\begin{tabular}{l l l r r r r}
\toprule
\textbf{Region} & \textbf{Category} & \textbf{Month} & \textbf{Revenue} & \textbf{Units} \\
\midrule
North America & Electronics & January & 7999.9 & 10\\
North America & Electronics & February & 2899.9 & 12\\
\bottomrule
\end{tabular}
\end{table}

\end{document}